\documentclass{article}
\usepackage{spconf,amsmath,graphicx,hyperref}
\usepackage{enumitem, bm, amssymb, booktabs, multirow, cite, setspace, etoolbox, xcolor, marvosym}

\title{Wavelet-Aware Anomaly Detection in Multi-Channel User Logs via Deviation Modulation and Resolution-Adaptive Attention}

\name{
	Kaichuan Kong\textsuperscript{1}\quad
	Dongjie Liu\textsuperscript{1*}\quad
	Xiaobo Jin\textsuperscript{2}\quad
	Shijie Xu\textsuperscript{1}\quad
	Guanggang Geng\textsuperscript{1}\quad
}

\address{
	$^{1}$College of Cyber Security, Jinan University, Guangzhou, China\\
	$^{2}$School of Advanced Technology, Xi’an Jiaotong-Liverpool University, Suzhou, China
}

\makeatletter
\patchcmd{\@maketitle}{\vskip 1.5em}{\vskip 1em}{}{}
\patchcmd{\@maketitle}{\@name \\ \@address}{\@name \\[-1em] \@address}{}{}
\patchcmd{\@maketitle}{\vskip 1.5em}{\vskip 1em}{}{}
\makeatother

\begin{document}

\maketitle
\begin{abstract}
Insider threat detection is a key challenge in enterprise security, relying on user activity logs that capture rich and complex behavioral patterns. These logs are often multi-channel, non-stationary, and anomalies are rare, making anomaly detection challenging. To address these issues, we propose a novel framework that integrates wavelet-aware modulation, multi-resolution wavelet decomposition, and resolution-adaptive attention for robust anomaly detection. Our approach first applies a deviation-aware modulation scheme to suppress routine behaviors while amplifying anomalous deviations. Next, discrete wavelet transform (DWT) decomposes the log signals into multi-resolution representations, capturing both long-term trends and short-term anomalies. Finally, a learnable attention mechanism dynamically reweights the most discriminative frequency bands for detection. On the CERT r4.2 benchmark, our approach consistently outperforms existing baselines in precision, recall, and F1 score across various time granularities and scenarios. 

\end{abstract}

\vspace{-1em}
\section{Introduction}
\label{sec:intro}
\vspace{-0.5em}
Insider threat detection is a critical security task in enterprise environments, where authenticated users may intentionally or unintentionally engage in malicious activities. To support such detection, user activity logs capturing access patterns, resource interactions, and system operations—offer rich behavioral signals~\cite{alzaabi2024review}. These logs are inherently multi-channel and time-evolving, reflecting both normal routines and potential threats~\cite{duan2025eagerlog,zhang2025scalalog,zhou2025logsi}.However, detecting anomalies in these logs remains highly challenging due to three key factors: (i) the multi-source and high-dimensional structure of the data; (ii) the rarity and subtlety of abnormal behaviors, often obscured by dominant routine patterns; and (iii) the non-stationary temporal dynamics that blend long-term trends with short-term bursts.

Previous studies have utilized statistical features, CNNs, and sequence models such as RNNs and Transformers for log modeling~\cite{he2021insider,xiao2024unveiling}. However, several limitations remain: Many models are sensitive to redundant operations; Imbalanced distributions lead to overfitting; and the cost of modeling long distances is generally high. In addition, frequency domain methods (e.g., FFT) can capture global changes~\cite{yang2024shortterm, kim2023anomaly}, but their coarse granularity hinders the detection of local anomalies.

To address these challenges, we propose a wavelet-aware, multi-resolution framework for detecting anomalies in multi-channel user activity logs. First, the raw logs are aggregated into a structured behavior matrix and modulated by a deviation-aware scheme that suppresses stable regular behaviors while emphasizing irregular deviations. Second, a multi-resolution representation is obtained via the discrete wavelet transform (DWT)\cite{nakamura2020time}, distinguishing stable long-term trends from transient bursts. Finally, a resolution-adaptive attention module adaptively reweights frequency components, enhancing the most informative scales for anomaly detection. Overall, our approach balances robustness and computational efficiency.

The principal contributions of this work are as follows:
\begin{itemize}[leftmargin=*, itemsep=1pt, parsep=1pt, topsep=1pt]

\item We propose a deviation-aware modulation mechanism that adaptively weights the behavior matrix to highlight anomalous deviations.

\item We integrate multi-resolution wavelet decomposition to extract frequency-aware temporal features from the behavior matrix.

\item We design a resolution-adaptive attention mechanism to adaptively calibrate multi-scale representations.

\item Experiments on the CERT benchmark show that our approach outperforms state-of-the-art baselines across a variety of scenarios and granularity levels.

\end{itemize}
    
\begin{figure*}[!t]
    \centering
    \includegraphics[width=0.88\textwidth]{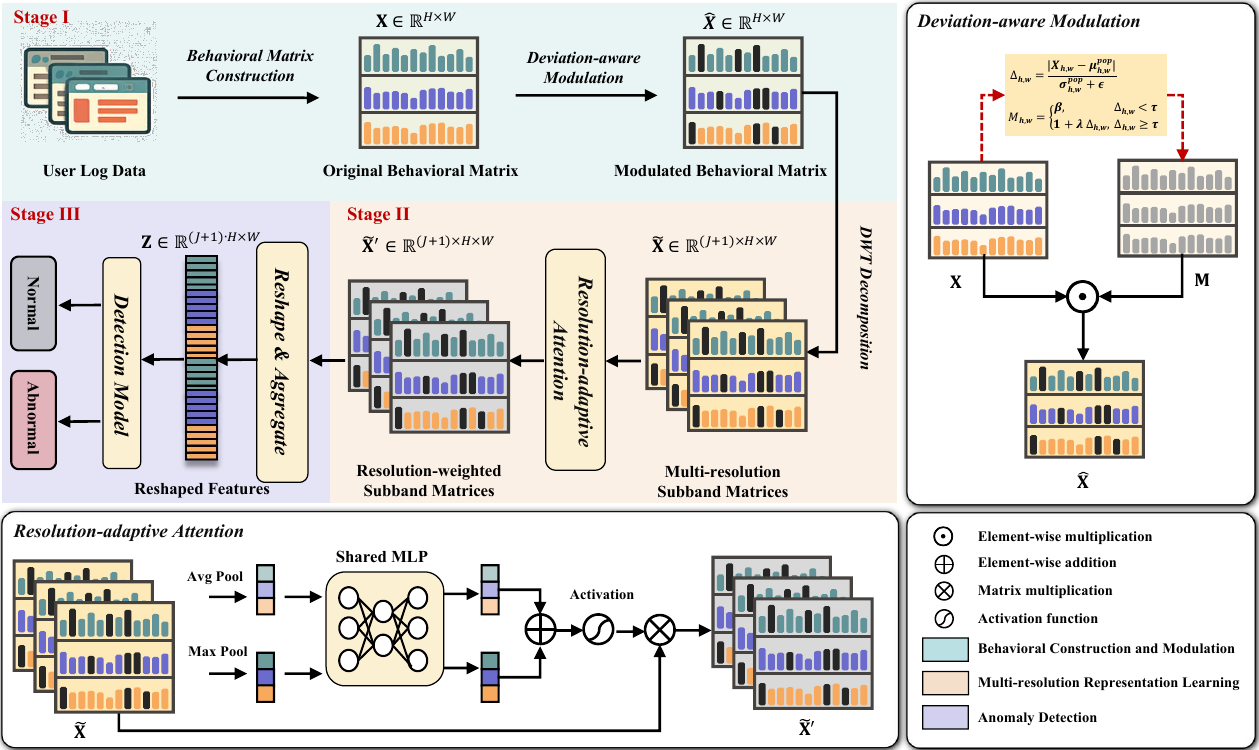}
    \caption{Overview of our framework. Raw user logs are converted into an action matrix $\mathbf{X}\!\in\!\mathbb{R}^{H\times W}$, which is then modulated by bias-aware weights $\mathbf{M}\!\in\!\mathbb{R}^{H\times W}$, resulting in a modulation matrix $\hat{\mathbf{X}}=\mathbf{X}\odot\mathbf{M}$ (right). The modulation matrix is decomposed into $J{+}1$ resolution subbands $\tilde{\mathbf{X}}\!\in\!\mathbb{R}^{(J{+}1)\times H\times W}$ using a DWT, where the weights are determined by a resolution-adaptive attention mechanism. This matrix, comprised of multiple resolution subbands, is then fed into the detection model to produce anomaly results.}
    \label{fig:framework}
\vspace{-1em}
\end{figure*}

\vspace{-1em}
\section{Related Work}
\label{sec:related}
\vspace{-0.5em}
Early log-based insider threat detection methods relied on classic machine learning models with artificial features. Techniques such as Log2Vec ~\cite{liu2019log2vec} and information gain (IG) and correlation-based feature selection (CFS) ~\cite{bin2022insider} were used to model underlying user behavior. Models such as random forest (RF), support vector machine (SVM), and extreme gradient boosting (XGBoost) demonstrated promising performance ~\cite{le2020analyzing}. In unsupervised settings, isolation forest (iForest), local outlier factor (LOF), and autoencoders ~\cite{bartoszewski2021anomaly, le2021anomaly, yousef2023machine} were used to detect behavioral deviations from audit logs.

To address temporal and semantic constraints, deep learning approaches such as attention-based long short-term memory (Att-LSTM)~\cite{he2021insider}, BERT (Bidirectional Encoder Representations from Transformers)-LSTM~\cite{huang2021itdbert}, and Transformer-based architectures~\cite{xiao2024unveiling} have been explored. Representation learning has also advanced through generative adversarial networks (GANs)~\cite{gayathri2024spcagan}, time-frequency analysis~\cite{budvzys2024deep}, and graph neural networks (GNNs)~\cite{roy2024graphch, cai2024lan} for user relationship modeling. Recently, large language models (LLMs) have achieved zero-shot/few-shot detection of log sequences through hinting~\cite{qi2023loggpt, liu2024interpretable} or fine-tuning~\cite{song2025confront}. 

In recent years, signal processing methods for audit log anomaly detection have emerged, leveraging wavelet transforms for time-frequency decomposition. For example, the discrete wavelet transform (DWT) has been used for anomaly scoring (\cite{feng2017wavelet}); capsule networks (CapsNet) with multi-level wavelet features have achieved high accuracy (\cite{randive2023mwcapsnet}); and denoising techniques have improved clustering-based models (\cite{kim2023anomaly}). These methods show promise, but typically use fixed basis functions and univariate signals, limiting their adaptability to dynamic multi-channel behavior.

\vspace{-1em}
\section{Methodology}
\label{sec:methodology}
\vspace{-0.5em}

Fig.~\ref{fig:framework} illustrates the overall framework. 
Our approach consists of two main stages: (i) transforming raw logs into structured and deviation-aware behavioral matrix, and (ii) learning multi-resolution representations that capture both stable routines and irregular anomalies. 
The following subsections detail each stage.

\vspace{-0.5em}
\subsection{Behavioral Construction and Modulation}
\vspace{-0.5em}
User activity logs are inherently sparse, irregular, and dominated by routine behaviors, making direct modeling ineffective. We first construct behavioral matrices to convert raw logs into temporal sequences, then apply deviation-aware modulation to suppress redundant patterns while amplifying informative anomalies.

\vspace{0.5em}
\noindent\textbf{Behavioral Matrix Construction.}
Event logs contain heterogeneous records (e.g., logon, file access, email), each with a timestamp.
Let $\mathcal{E}$ be the set of recorded events, where each event $e\in\mathcal{E}$ is described by a type $e.\text{type}\in\mathcal{H}$ and timestamp $e.\text{time}$.
Here $\mathcal{H}$ is the set of behavior categories with cardinality $H{=}\,|\mathcal{H}|$.
We divide the timeline into $W$ equal-width, non-overlapping intervals $\{w_1,\dots,w_W\}$ (e.g., hourly slots).
The behavior matrix $\mathbf{X} \!\in\! \mathbb{R}^{H \times W}$ is:
\begin{equation}
\vspace{-0.2em}
X_{h,w} = \sum_{e \in \mathcal{E}} \mathbb{I}\{e.\text{type}=h,~e.\text{time}\in w\},
\vspace{-0.3em}
\end{equation}
where $\mathbb{I}\{\cdot\}$ is the indicator function. This yields a structured representation with rows for behavior types and columns for temporal intervals.

\vspace{0.5em}
\noindent\textbf{Deviation-aware Modulation.}
Routine activities (e.g., office-hour logins) often dominate $\mathbf{X}$ and overshadow rare but informative anomalies.
To provide a population baseline for each coordinate, we compute $\mu_{h,w}$ and $\sigma_{h,w}$ as the mean and standard deviation of $X^{(n)}_{h,w}$ across \emph{normal training samples only}, fixed after training and used at inference. The standardized deviation score is defined as:
\begin{equation}
\vspace{-0.2em}
\Delta_{h,w} = \frac{|X_{h,w}-\mu_{h,w}|}{\sigma_{h,w}+\epsilon},
\vspace{-0.3em}
\end{equation}
where $\epsilon>0$ is a small constant for numerical stability.
We then compute the weight matrix $\mathbf{M}$:
\begin{equation}
\vspace{-0.3em}
M_{h,w} =
\begin{cases}
\beta, & \Delta_{h,w} < \tau, \\
1+\lambda \Delta_{h,w}, & \Delta_{h,w} \ge \tau,
\end{cases}
\vspace{-0.3em}
\end{equation}
where $\beta\in(0,1)$ suppresses stable behaviors, $\lambda>0$ amplifies deviations, and $\tau\ge 0$ is tuned on validation data.
The modulated matrix is:
\begin{equation}
\vspace{-0.2em}
\hat{\mathbf{X}} = \mathbf{X} \odot \mathbf{M},
\vspace{-0.3em}
\end{equation}
where $\odot$ is the Hadamard product. This reweighting suppresses frequent yet uninformative events and sharpens rare deviations, yielding a clearer representation for downstream anomaly analysis.

\vspace{-0.5em}
\subsection{Multi-resolution Representation Learning}
\vspace{-0.5em}
User behaviors unfold across multiple temporal scales, from long-term work routines to sudden abnormal bursts. To capture both stable patterns and transient deviations, we perform multi-resolution decomposition with wavelets and further design an attention mechanism to adaptively emphasize the most informative scales.

\vspace{0.5em}
\noindent\textbf{Wavelet-based Multi-resolution Decomposition.}
User behaviors are rarely confined to a single temporal scale. 
For example, weekly email cycles coexist with sudden bursts of abnormal data transfer. 
To capture both coarse regularities and fine-grained irregularities, we employ the discrete wavelet transform (DWT)~\cite{nakamura2020time}, a classical tool in signal processing for multi-resolution analysis.  

For each sequence $\hat{X}_{h,:}\in\mathbb{R}^W$, a $J$-level DWT yields:
\begin{equation}
\vspace{-0.2em}
\hat{X}_{h,:} = A_J^{(h)} + \sum_{j=1}^J D_j^{(h)},\quad h = 1,2,\cdots,H,
\vspace{-0.2em}
\end{equation}
where $A_J^{(h)}$ captures the low-frequency approximation (long-term trend), while $\{D_j^{(h)}\}$ are detail coefficients corresponding to high-frequency fluctuations at different scales.  

Since these components have different lengths, we interpolate them back to the original resolution $W$ and stack them:
\begin{equation}
\vspace{-0.2em}
\tilde{\mathbf{X}}\in\mathbb{R}^{(J+1)\times H \times W}.
\vspace{-0.3em}
\end{equation}

This decomposition enriches the matrix representation: $A_J^{(h)}$ reflects stable user routines, while $D_j^{(h)}$ highlights transient bursts that may correspond to security violations.

\vspace{0.5em}
\noindent\textbf{Resolution-adaptive Attention Aggregation.}
Not all decomposed components are equally useful: low-frequency approximations may dilute anomalies, while certain high-frequency bands may contain mostly noise. 
To adaptively focus on informative scales, we design a channel attention mechanism across resolutions.  
After reweighting, all components are concatenated into a unified embedding, which compactly represents multi-resolution user behavior.

For each resolution $c$, we extract two descriptors:
\begin{equation}
\vspace{-0.2em}
z_c^{\text{avg}}=\frac{1}{HW}\sum_{h,w}\tilde{\mathbf{X}}[c,h,w], \quad
z_c^{\text{max}}=\max_{h,w}\tilde{\mathbf{X}}[c,h,w].
\vspace{-0.2em}
\end{equation}
The descriptors are concatenated and transformed by a two-layer perceptron:
\begin{equation}
\vspace{-0.2em}
s=\sigma\!\left(W_2 \,\delta(W_1 [z^{\text{avg}}\Vert z^{\text{max}}])\right), \quad s\in\mathbb{R}^{J+1},
\vspace{-0.2em}
\end{equation}
where $\delta(\cdot)$ is ReLU and $\sigma(\cdot)$ is sigmoid.  
The weight $s_c$ determines the importance of resolution $c$, and the reweighted tensor is:
\begin{equation}
\vspace{-0.2em}
\tilde{\mathbf{X}}'[c,h,w]=s_c \cdot \tilde{\mathbf{X}}[c,h,w].
\vspace{-0.2em}
\end{equation}

\vspace{-0.5em}
\subsection{Anomaly Detection}
\vspace{-0.5em}
The reweighted components are concatenated and reshaped into the embedding:
\begin{equation}
\vspace{-0.2em}
\mathbf{Z}=\text{Reshape}(\tilde{\mathbf{X}}')\in\mathbb{R}^{(J+1)\cdot H \times W},
\vspace{-0.2em}
\end{equation}
which jointly captures semantic behavior types and frequency-aware significance. 

Given an anomaly detector $f(\cdot)$, the prediction is obtained as:
\begin{equation}
\vspace{-0.2em}
\hat{y}=f(\mathbf{Z}), \quad \hat{y}\in\{\text{Normal}, \text{Abnormal}\}.
\vspace{-0.2em}
\end{equation}

\begin{table*}[htbp]
\centering
\renewcommand{\arraystretch}{0.8}   
\setlength{\tabcolsep}{3pt} 
\caption{Comparison of detection performance of multiple algorithms in different scenarios}
\label{tab:comparison}
\begin{tabular}{c c  ccc  | ccc | ccc  | ccc | ccc }
\toprule
{\multirow{2}{*}{\textbf{Scenario}}}  & {\multirow{2}{*}{\textbf{Granularity}}} 
& \multicolumn{3}{c}{\textbf{DWT-OCSVM}} 
& \multicolumn{3}{c}{\textbf{MWCapsNet}} 
& \multicolumn{3}{c}{\textbf{CATE}}
& \multicolumn{3}{c}{\textbf{ITDLM}} 
& \multicolumn{3}{c}{\textbf{Ours}} \\
\cmidrule{3-5} \cmidrule{6-8} \cmidrule{9-11} \cmidrule{12-14} \cmidrule{15-17}
& & Pre & Rec & F1 & Pre & Rec & F1 & Pre & Rec & F1 & Pre & Rec & F1 & Pre & Rec & F1 \\
\midrule
{\multirow{3}{*}{\textbf{Scenario 1}}}  
& 24h  & 0.88 & 1.00 & 0.93 & 0.93 & 1.00 & 0.96 & 0.99 & 0.99 & 0.99 & 0.89 & 0.92 & 0.90 & 0.98 & 0.99 & 0.98 \\
& 72h  & 0.85 & 1.00 & 0.92 & 0.99 & 0.99 & 0.99 & 0.98 & 0.99 & 0.99 & 0.91 & 0.93 & 0.93 & 0.97 & 0.98 & 0.97 \\
& 168h & 0.80 & 1.00 & 0.89 & 0.99 & 0.99 & 0.99 & 0.50 & 0.98 & 0.67 & 0.94 & 0.98 & 0.96 & 0.96 & 0.98 & 0.97 \\
\midrule
{\multirow{3}{*}{\textbf{Scenario 2}}} 
& 24h  & 0.90 & 0.59 & 0.72 & 0.79 & 0.80 & 0.80 & 0.77 & 0.87 & 0.82 & 0.83 & 0.87 & 0.83 & 0.96 & 0.95 & 0.96 \\
& 72h  & 0.85 & 0.77 & 0.81 & 0.82 & 0.95 & 0.88 & 0.76 & 0.93 & 0.84 & 0.86 & 0.89 & 0.90 & 0.96 & 0.97 & 0.96 \\
& 168h & 0.74 & 0.93 & 0.83 & 0.80 & 0.97 & 0.88 & 0.61 & 0.96 & 0.75 & 0.92 & 0.93 & 0.93 & 0.97 & 0.98 & 0.98 \\
\midrule
{\multirow{3}{*}{\textbf{Scenario 3}}} 
& 24h  & 0.89 & 1.00 & 0.94 & 0.93 & 0.99 & 0.96 & 0.98 & 0.99 & 0.98 & 0.87 & 0.89 & 0.89 & 0.98 & 0.98 & 0.98 \\
& 72h  & 0.91 & 1.00 & 0.95 & 0.99 & 0.99 & 0.99 & 0.99 & 0.99 & 0.99 & 0.89 & 0.91 & 0.90 & 0.97 & 0.98 & 0.97 \\
& 168h & 0.86 & 1.00 & 0.92 & 0.97 & 0.99 & 0.98 & 0.99 & 0.99 & 0.99 & 0.94 & 0.96 & 0.95 & 0.96 & 0.97 & 0.97 \\
\midrule
\textbf{Avg.} & -- 
& 0.85 & 0.93 & 0.89 
& 0.93 & 0.95 & 0.95 
& 0.84 & 0.95 & 0.90 
& 0.90 & 0.93 & 0.92 
& \textbf{0.97} & \textbf{0.98} & \textbf{0.97} \\
\textbf{Rank} & -- 
& -- & -- & 5 
& -- & -- & 2 
& -- & -- & 4 
& -- & -- & 3 
& -- & -- & \textbf{1} \\
\bottomrule
\end{tabular}
\vspace{-0.5cm}  
\end{table*}

    \begin{table}[htbp]
        \centering
        \vspace{-0.3cm}  
        \renewcommand{\arraystretch}{0.8}  
        \caption{Comparison of detection models}
        \label{tab:model}
        \begin{tabular}{lccc}
        \toprule
        \textbf{Model} & \textbf{Pre} & \textbf{Rec} & \textbf{F1} \\
        \midrule
        AE          & 0.49 & 0.98 & 0.66 \\
        IForest     & 0.83 & 0.66 & 0.73 \\
        XGBoost     & 0.98 & \textbf{1.00} & \textbf{0.99} \\
        Transformer & 0.87 & 0.93 & 0.90 \\
        TCN         & \textbf{1.00} & 0.98 & \textbf{0.99} \\
        \bottomrule
        \end{tabular}
        \vspace{-0.5cm}  
    \end{table}
    
	\vspace{-1em}
	\section{Experiments}
	\label{sec:experiments}
	\vspace{-0.5em}
    
To evaluate the effectiveness of our model, we conduct comparative experiments with state-of-the-art baselines on widely used benchmark datasets.
\vspace{-1em}
\subsection{Experimental Setup}
\vspace{-0.5em}
\noindent\textbf{Datasets and Baselines.} 
CERT r4.2 ~\cite{lindauer2020insider} is a public benchmark that contains anomaly labels for user activity logs in three scenarios: (i) after-hours activity leading to data breaches; (ii) data theft due to job hunting; and (iii) insider abuse by disgruntled administrators. The benchmark includes signal decomposition methods (DWT-OCSVM~\cite{kim2023anomaly}, MWCapsNet~\cite{randive2023mwcapsnet}), a Transformer-based model (CATE~\cite{xiao2024unveiling}), and an LLM fine-tuning method (ITDLM~\cite{song2025confront}).

\noindent\textbf{Implementation Details.} 
Logs are segmented into \emph{behavior matrices} with window sizes of 24 hours (daily), 72 hours (three-day), and 168 hours (weekly), with a step size of 24 hours. Any unusual event within a window is marked as an anomaly. For each scenario, we randomly select three users and report their average performance to mitigate user-specific biases. All models are implemented in PyTorch and trained on four NVIDIA L20 GPUs. To verify generalizability, we instantiate five detectors: three classical methods (Autoencoder~\cite{singh2024network}, IForest, and XGBoost~\cite{le2020analyzing}) and two deep models (Transformer~\cite{vaswani2017attention} and TCN~\cite{fan2023parallel}).
We report \textbf{Precision}, \textbf{Recall}, and \textbf{F1} scores.

\vspace{-1em}
\subsection{Main Results and Analysis}
\vspace{-0.5em}

Table~\ref{tab:comparison} reports detection results in three experimental scenarios, each of which is further evaluated at multiple temporal granularities (24-hour, 72-hour, and 168-hour windows). Across all settings, our proposed method achieves the most balanced and robust performance, with average precision, recall, and F1 of 0.97, 0.98, and 0.97, respectively. Compared to the traditional DWT-OCSVM and recent baselines such as MWCapsNet, CATE, and ITDLM, our method maintains consistently high precision and recall across various scenarios and granularities. In particular, while some baselines tend to increase precision or recall depending on window size, our method achieves stable improvements on all three metrics and achieves the best overall ranking. These results confirm the effectiveness of our framework in capturing multi-scale user behavior dynamics and ensuring reliable anomaly detection in a variety of usage scenarios.

\vspace{-1em}
\subsection{Detection Model Selection}
\vspace{-0.5em}
To test the applicability of different detectors with our representation, we conducted a model selection study using a representative user in \textbf{Scenario~2}. As shown in Table~\ref{tab:model}, traditional unsupervised methods such as autoencoders and IForest performed moderately well, while XGBoost and deep sequence models (Transformer, TCN) consistently achieved high F1 scores. \textbf{TCN and XGBoost, in particular, achieved F1 scores of 0.99}, highlighting their effectiveness in capturing temporal dependencies and discriminative patterns. This experiment validated the compatibility of our framework with various detection paradigms, and we will use the best-performing models in subsequent evaluations.

\vspace{-1em}
\subsection{Ablation Study}
\vspace{-0.5em}
Table~\ref{tab:ablation} shows the ablation results. The full model achieves the best performance (F1=0.99). Removing bias modulation reduces recall (1.00→0.90), highlighting its role in highlighting anomalous behavior. Removing the discrete wavelet transform reduces F1 to 0.94, demonstrating the importance of multiresolution decomposition in capturing temporal dynamics. Finally, removing the attention mechanism reduces F1 further to 0.93, confirming the effectiveness of adaptively adjusting resolution weights. These results confirm that each component contributes and that their combination yields the best performance.

\begin{table}[tbp]
    \centering
    \vspace{-0.3cm}  
    \renewcommand{\arraystretch}{0.8}
    \caption{Ablation study on components}
    \label{tab:ablation}
    \begin{tabular}{lccc}
    \toprule
    \textbf{Model} & \textbf{Pre} & \textbf{Rec} & \textbf{F1} \\
    \midrule
    Ours (Full Model)        & \textbf{0.98} & \textbf{1.00} & \textbf{0.99} \\ 
    w/o Deviation Modulation & 0.96 & 0.90 & 0.92 \\ 
    w/o DWT                  & 0.93 & 0.94 & 0.94 \\
    w/o Attention            & 0.94 & 0.92 & 0.93 \\
    \bottomrule
    \end{tabular}
    \vspace{-0.5cm}  
\end{table}

\vspace{-1em}
\section{Conclusion}
\label{sec:conclusion}
\vspace{-0.6em}

In this study, We propose a wavelet-aware multi-resolution framework for log anomaly detection on multi-channel user activity logs. By converting raw logs into a structured behavior matrix, decomposing it via discrete wavelet transform (DWT), and applying a resolution-adaptive attention mechanism, our approach effectively distinguishes regular operations from anomalous deviations. Experimental results show that our approach consistently outperforms state-of-the-art machine learning and deep learning baselines in terms of precision, recall, and F1 score.  In the future, we plan to extend our framework to larger and more diverse log datasets (e.g., cloud infrastructure logs) or enhance streaming log analysis frameworks through dynamic threshold adjustment.
    
\vfill\pagebreak
\label{sec:refs}
{ 
    \small
    \begin{spacing}{0.85}
        \bibliographystyle{IEEEbib}
        \bibliography{ref}
    \end{spacing}
}
	
\end{document}